\documentclass[letterpaper, 10 pt, conference]{ieeeconf}  

\usepackage{cite}
\usepackage{amsmath,amssymb,amsfonts}
\usepackage{algorithm}
\usepackage{algorithmic}
\usepackage{graphicx}
\usepackage{textcomp}
\usepackage{xcolor}
\usepackage{booktabs}
\usepackage{hyperref}
\usepackage{siunitx}

\IEEEoverridecommandlockouts                              

\title{\LARGE \bf
Realizing Text-Driven Motion Generation on NAO Robot: A Reinforcement Learning-Optimized Control Pipeline
}

\author{Zihan Xu\textsuperscript{\dag}, Mengxian Hu\textsuperscript{\dag}, Kaiyan Xiao, Qin Fang, Chengju Liu\textsuperscript{*} and Qijun Chen, \emph{Senior Member, IEEE}
\thanks{{\dag} are equal contributors, {*} is the corresponding author.}%
\thanks{Authors are with the College of Electronic and Information Engineering, Tongji University, Shanghai, China. \{xuzihan, humengxian, xiaokaiyan, tongji\_fq, liuchengju, qjchen\}@tongji.edu.cn}%
}

\begin{document}

\maketitle
\thispagestyle{empty}
\pagestyle{empty}

\begin{abstract}

Human motion retargeting for humanoid robots, transferring human motion data to robots for imitation, presents significant challenges but offers considerable potential for real-world applications. Traditionally, this process relies on human demonstrations captured through pose estimation or motion capture systems. In this paper, we explore a text-driven approach to mapping human motion to humanoids. To address the inherent discrepancies between the generated motion representations and the kinematic constraints of humanoid robots, we propose an angle signal network based on norm-position and rotation loss (NPR Loss). It generates joint angles, which serve as inputs to a reinforcement learning-based whole-body joint motion control policy. The policy ensures tracking of the generated motions while maintaining the robot's stability during execution. Our experimental results demonstrate the efficacy of this approach, successfully transferring text-driven human motion to a real humanoid robot NAO.

\end{abstract}


\section{Introduction}
Humanoid robots have long been recognized for their potential to mimic human actions due to their anthropomorphic structure. While their design naturally lends itself to imitation, achieving the seamless transfer of human motion to robots remains a complex challenge\cite{koenemann2014real}. Recent advancements in text-driven diffusion models offer a promising solution by generating human motion from textual descriptions\cite{tevet2022motionclip,petrovich2021action,Zhang2022}, paving the way for more flexible and intuitive motion generation. Applying these human motions to humanoid robots requires overcoming the structural and kinematic discrepancies between humans and humanoids. 

Directly applying human motion data to robotic control is far from straightforward while matching the end-effector pose with the target's pose yields higher similarity and practical applicability. From this perspective, end-effector tracking is not only more effective but also more aligned with real-world applications. To achieve reliable end-effector tracking, human motion data must first be processed in a way that is compatible with the robot's kinematics. Traditional human pose estimation methods typically infer joint orientations from captured human poses, which are then used to compute robot joint angles via inverse kinematics\cite{li2021hybrik}. Alternatively, using parametric human models to parameterize the robot's structure provides more consistent motion transfer, reformulating the imitation task as a keypoint tracking problem\cite{he2024learning}.
\begin{figure}[!t]
    \centering
    \includegraphics[width=\linewidth]{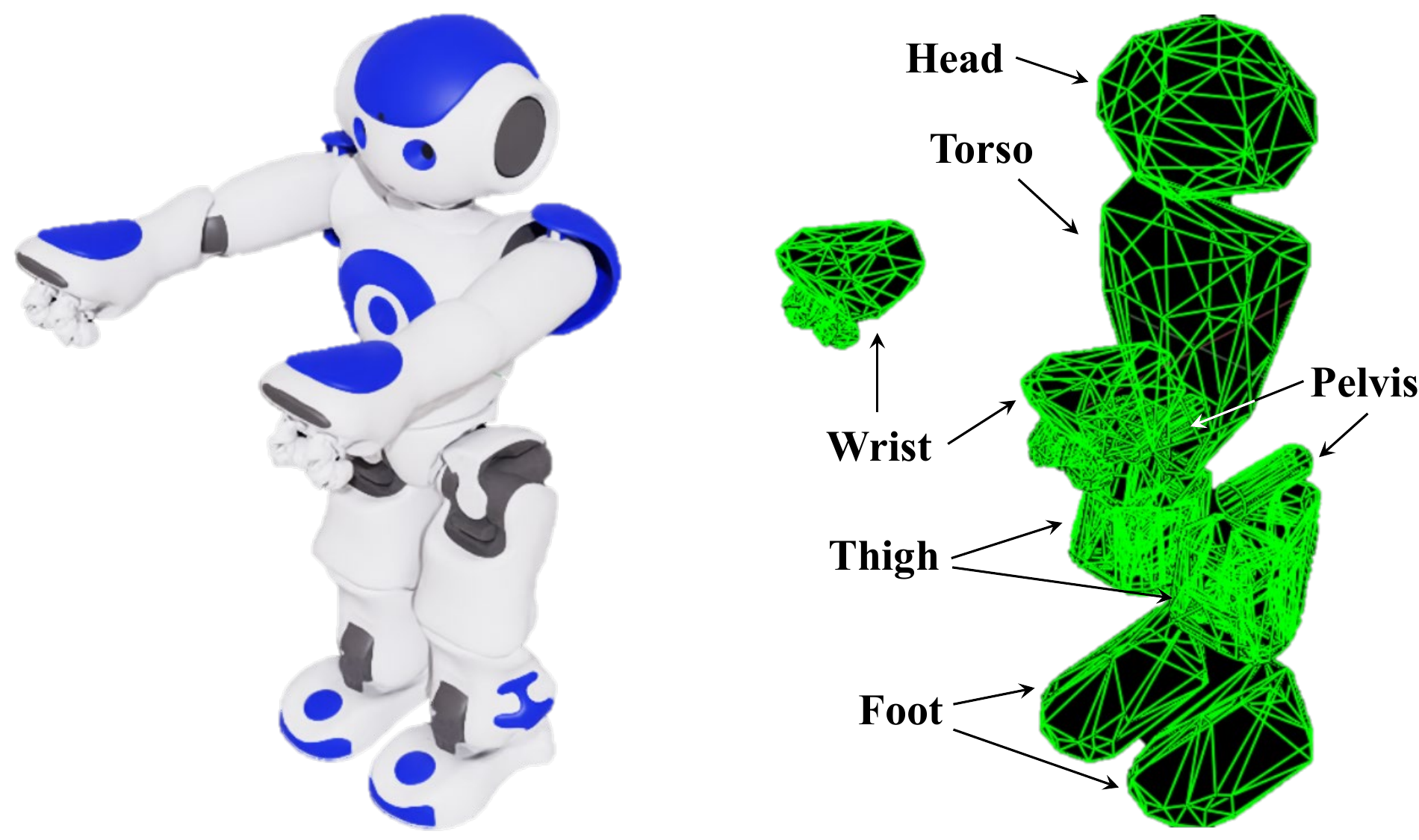}
    \caption{Visualization of NAO robot: Appearance (left) and collision model (right). Except for the thigh link, which uses the convex decomposition method to generate the colliders, all other links adopt the convex hull method, generating colliders with up to 64 faces.}
    \label{fig: nao colliders}
\end{figure}


From the robot's perspective, human motion provides a high-dimensional solution space $S \subset \mathbb{R}^n$. The goal of retargeting human motion to a humanoid is to identify a subset $S' \subseteq S$ that satisfies a set of predefined constraints or conditions. Fundamentally, we aim to encode human movements into a mathematical representation that captures essential parameters: the relative position and orientation of the end-effector with respect to the root joint of the articulated chain. These relative parameters are dimensionless, ensuring that they can generalize across both human and humanoid motions. This parameterization allows us to decode the posture regardless of whether the source is human or humanoid motion.

\begin{figure*}[!t]
    \centering
    \includegraphics[width=.9\linewidth]{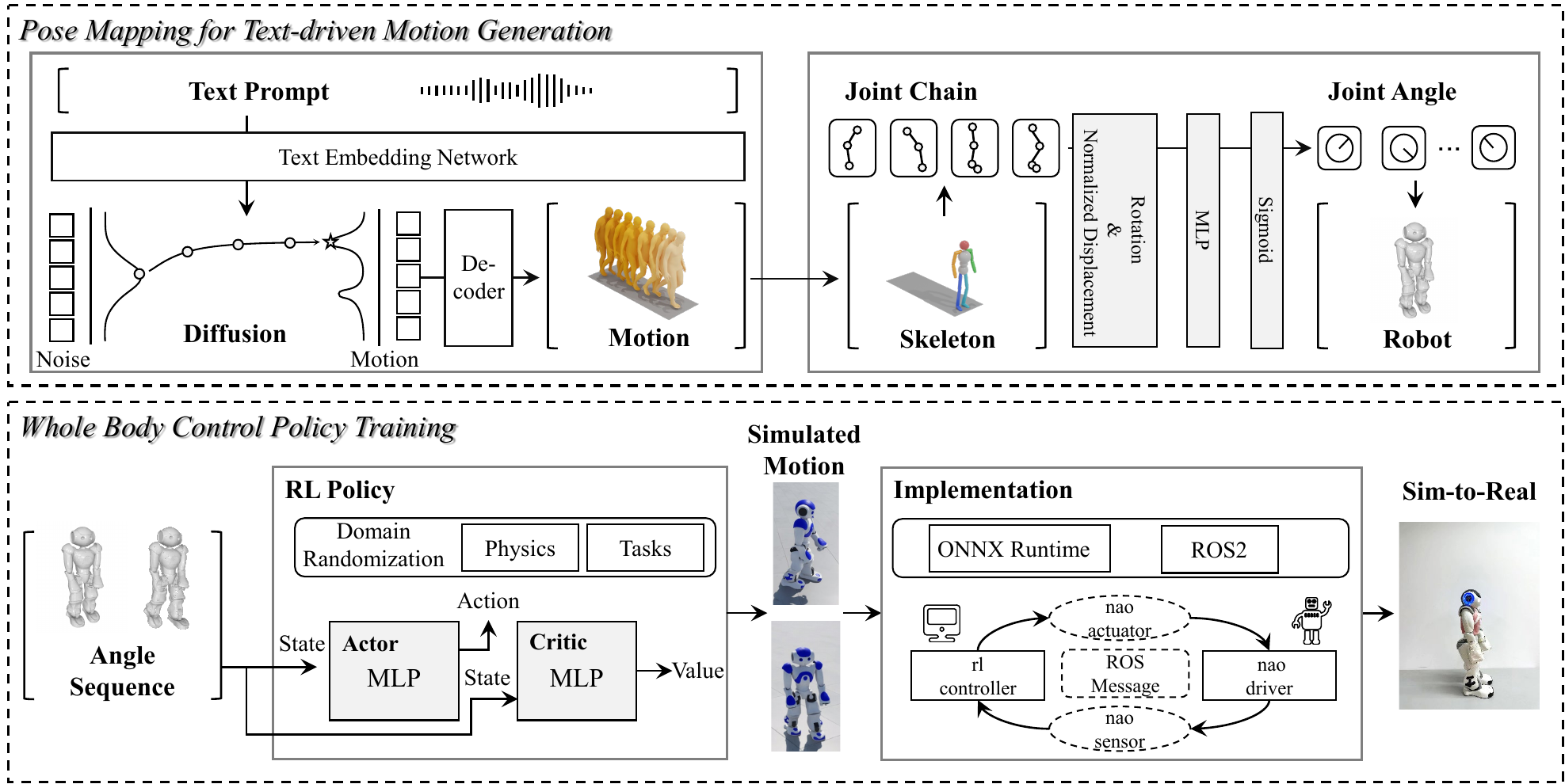}
    \caption{Overview of mapping text-driven human motion to humanoids: (a) \textbf{Pose Mapping} (Section~\ref{sec: Pose Mapping for Text-Generated Motion}): Mapping  the text-driven human motion generation to the robot joint angles by angle signal network. (b) \textbf{Whole Body Control} (Section~\ref{sec: Whole Body Control Policy Training}): Training the reinforcement-learning based controller and implement the policy using ONNX Runtime and run the system in ROS2 framework.}
    \label{fig:system diagram}
\end{figure*}

Instead of directly transferring human motion to humanoid robots, we employ a norm-based description of motion to define as the basis for our normalized mapping loss function, as discussed in Section~\ref{sec: Normalized mapping of humans to humanoids}. This description encompasses both norm-position and rotational components. For human movements, we compute the normalized parameters. For humanoid robots, the objective is to determine the joint configurations that best replicate the posture under the described parameters. This approach not only enhances the fidelity of imitation but also ensures that the robot's motion closely adheres to the intended end-effector trajectory, making the solution adaptable and scalable to different motion tasks.


In this paper, we use text-driven diffusion model to generate reference joint commands through angle signal net and optimized the joint commands to reliable joint output based on the IsaacLab reinforcement learning frameworks as illustrated in Fig.~{\ref{fig:system diagram}}. The whole-body control is achieved and applied in a real-world robot. The main contributions include: (\textbf{I}) A text-driven diffusion model was leveraged to generate motion reference sequences, and employed an angle signal network based on NPR Loss to achieve precise pose mapping. (\textbf{II}) The detailed NAO robot simulation model, derived from its URDF, was developed, with its collision volumes fine-tuned to ensure full compatibility with the IsaacSim 4.5 platform. Furthermore, we identified practical joint actuator parameters tailored for real-world NAO applications. (\textbf{III}) The robot motion control system was designed using the ROS2 framework, enabling the successful deployment of reinforcement learning policies for controlling a physical NAO robot. All components of this work, including the simulation model and control system, are fully open-sourced to facilitate further research and development in the field.

\section{Related Works}

Human motion imitation by humanoid robots has been a longstanding research focus, aiming to enable robots to perform tasks in a human-like manner. Previous works often rely on human pose estimation or motion capture data to generate control commands for robots\cite{mukherjee2015inverse,otani2017adaptive,viceconte2022adherent}. Khalil \textit{et al.} \cite{khalil2021human} employed methods based on inverse kinematics to compute joint angles from estimated human poses and apply to robots. These techniques primarily focus on upper-body motion transfer, neglecting leg movements and limiting the diversity of imitated actions. Directly mapping human joint angles to robot joints is challenging due to differences in kinematic structures and degrees of freedom (DoFs) between humans and robots.
To address these challenges, some researchers have parameterized the robot's structure using human models\cite{he2024learning}, defining keypoint correspondences to frame the imitation task as a keypoint tracking problem\cite{khansari2011learning,holden2016deep}. While this approach provides a structured framework that efficiently aligns human and robot kinematics, it still depends heavily on human-provided motion data, limiting the robot's ability to perform novel or unanticipated tasks without additional human input.

Text-driven motion generation has emerged as a promising solution to reduce reliance on human motion data by allowing robots to generate motions based on natural language descriptions. Early discussions focused on variational autoencoders (VAEs) \cite{Guo2020, Petrovich2021, Guo2022} and generative adversarial networks (GANs) \cite{Wang2020, Cai2018}. The former imposes a KL-divergence constraint on the distribution of latent representations and samples from a standard normal distribution, but often yields suboptimal results. The latter avoids explicit modeling of the target distribution through adversarial training, achieving good generative performance, though adversarial training is unstable and prone to mode collapse. Recent advancements include motion diffusion models \cite{Zhang2022, Chen2023, Jin2023}, which convert data into a standard normal distribution by progressively adding noise, and use multiple reverse diffusion steps to recover motion representations during inference. Compared to traditional single-step inference methods, the diffusion framework exhibits higher controllability and generative capability. Early related works, such as Motiondiffuse \cite{Zhang2022} and MDM \cite{Tevet2023}, demonstrated the feasibility of the motion diffusion framework. Subsequently, MLD \cite{Chen2023} and M2DM \cite{Kong2023} extended the diffusion process to the motion latent space, improving inference speed. The latest progress, MLCT \cite{hu2024efficient}, precomputes diffusion trajectories during the training phase, achieving high-quality, controllable motion generation with lower training and inference costs. These developments provide convenient data tools for imitation learning in humanoid robotics.

\begin{figure}[!t]
    \centering
    \includegraphics[width=\linewidth]{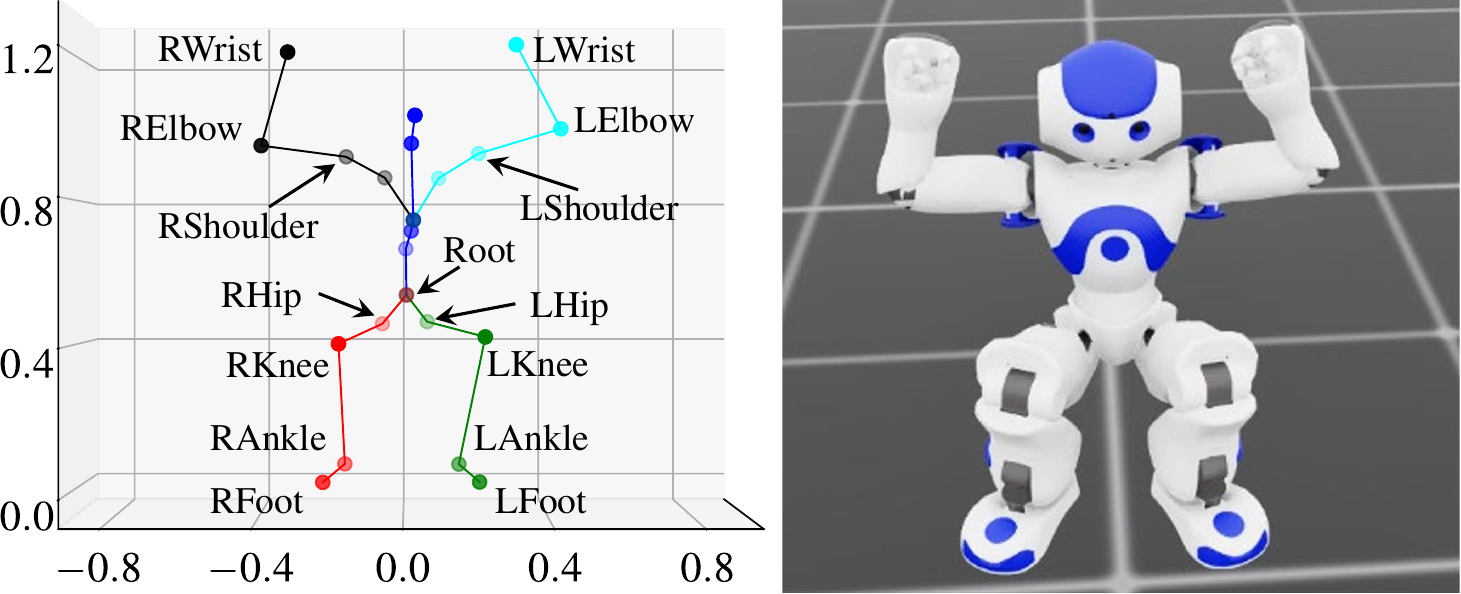}
    \caption{Skeletal keypoints derived from human motion data and mapped joints for NAO.}
    \label{fig: skeleton point}
\end{figure}

Building upon these developments, the subsequent challenge lies in ensuring that humanoid robots can execute the generated motions with stability in real-world environments. Reinforcement learning (RL) provides a framework for robots to learn complex motor skills through trial and error, control policies based on feedback from the environment\cite{peng2018deepmimic} and enables robots to adapt to dynamic environments\cite{tan2018sim}.
RL-based approaches have demonstrated remarkable success in bipedal locomotion, enabling robots to learn efficient walking and running strategies\cite{li2024ai}, recover from external disturbances\cite{castillo2021robust}, and generalize across different terrains and environments\cite{li2025impedance}.
A crucial aspect of deploying RL-trained policies in real-world settings is the sim-to-real transfer, which addresses the discrepancies between simulation and physical execution. Various techniques, such as domain randomization and adaptation strategies, have been proposed to bridge this gap\cite{zhao2020sim,hwangbo2019learning}. These approaches ensure that policies learned in simulation can generalize effectively to unstructured and unpredictable real-world scenarios. Benefiting from the development of large-scale high-performance simulation frameworks\cite{mittal2023orbit}, we can leverage GPU acceleration to enable massively parallelized training of robot policies. By integrating RL with text-driven motion generation, robots can learn to execute a wide range of motions specified by natural language, adapting them to their unique kinematic structures and environmental constraints.

\section{Pose Mapping for Text-generated Motion}
\label{sec: Pose Mapping for Text-Generated Motion}
\subsection{Synthesizing Text-driven Human Motion}
We focus on text-driven humanoid robot imitation strategies to assess the potential of imitation learning in human-robot interaction scenarios. To enable robust motion imitation, diverse and text-driven human motion sequences are the data cornerstone. Relying solely on a single motion capture dataset for imitation is suboptimal due to the high cost of motion capture equipment, which restricts sample diversity, and the semantic ambiguity arising from varying natural language descriptions of the same motion. To address these issues, we introduce the conditional generative model that estimates the human motion distribution and synthesize samples based on fine-grained semantic parsing.

\begin{figure}[!t]
    \centering
    \includegraphics[width=\linewidth]{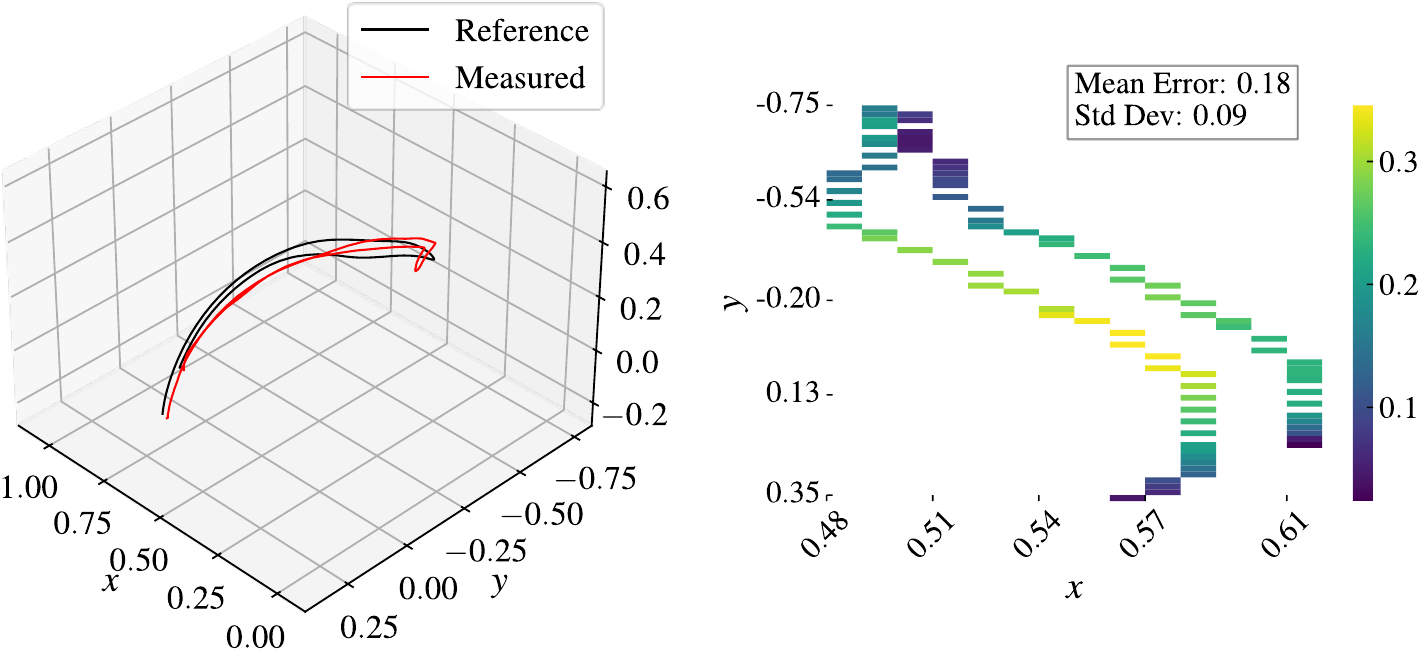}
    \caption{Comparison of reference and measured spatial trajectories for a waving motion, accompanied by a heatmap visualization of error distribution with mean and standard deviation.}
    \label{fig:wave hand trajectory}
\end{figure}

Motion diffusion models are a typical class of generative models with powerful distribution matching properties. It gradually injects noise into the human motion representation through a Gaussian perturbation kernel until the perturbed state approximates the standard normal distribution. Recent work indicates that the forward diffusion process can be described as the stochastic partial differential equation,
\begin{equation}\label{equ1}
    d x_t = f(t)x_tdt + g(t) dw_t
\end{equation}
where $t\in [\epsilon, T]$, $\epsilon$ and $T$ are the fixed positive constant, $w_{t}$ denotes the standard Brownian motion, $f$ and $g$ are the drift and diffusion coefficients respectively. The perturbation parameters are related as follows,
\begin{equation}\label{equ2}
    f(t) = \frac{d \log \alpha_t}{dt},\quad g^2(t)=\frac{d\sigma_t^2}{dt} - 2\frac{d\log\alpha_t}{dt}\sigma_t^2
\end{equation}
where $\alpha_t$ and $\sigma_t$ are noise schedules.

It is further demonstrated that the inverse diffusion process of Eq. \eqref{equ1} is approximated in solving the \textit{probabilistic flow ordinary differential equation}:
\begin{equation}\label{equ3}
    dx_t = [f(t)x_t - \frac{1}{2}g^2(t)\nabla_{x_t} \log p(x_t)]dt
\end{equation}
where $\nabla_x \log p(x_t)$ is named the \textit{score function}, fitted with a neural network $\mathcal{F}$. 
The traditional diffusion approach utilizes higher-order ODE numerical solvers to iteratively solve Eq. \eqref{equ3}, which involves expensive time and computational costs. To avoid such efficiency limitations, we utilize the motion consistency training framework, which precomputes the diffusion trajectories in the training phase to enable large-scale skip-step sampling in the inference phase.

Specifically, given a textual instruction $\mathcal{I}$, the CLIP model is employed as the textual feature extractor to parse fine-grained textual semantics.
Then, $x_T$ is sampled from the prior distribution $\mathcal{N}(0, I)$ and solved with reference the text embedding features following pre-computed reverse diffusion trajectories from Eq. \eqref{equ3} up to the motion lantent representation $x_\epsilon$. 
It's transformed with the motion decoder $\mathcal{D}$ into the human motion sequence $x=\mathcal{D}(x_\epsilon)\in \mathcal{R}^{f\times 22 \times 3}$ matching the instruction semantics, which contains the 3D coordinates of the 22 joints of $f$ motion frames.
The generated human pose data follows the format of the HumanML3D\cite{Guo_2022_CVPR} dataset.

\floatname{algorithm}{Procedure}
\begin{algorithm}[!t]
\caption{Computing Rotation Matrix Between Two Vectors}
\label{proc:rotation_matrix}
\begin{algorithmic}[1]
\REQUIRE $\boldsymbol{v}_1, \boldsymbol{v}_2 \in \mathbb{R}^3$ (non-zero vectors)
\ENSURE  $3 \times 3$ rotation matrix $\boldsymbol{R}$
\STATE Normalize input vectors: $ \hat{\boldsymbol{v}}_1 = \frac{\boldsymbol{v}_1}{\|\boldsymbol{v}_1\|}, \quad \hat{\boldsymbol{v}}_2 = \frac{\boldsymbol{v}_2}{\|\boldsymbol{v}_2\|} $
\STATE Compute rotation axis: $ \boldsymbol{k} = \hat{\boldsymbol{v}}_1 \times \hat{\boldsymbol{v}}_2 $
\STATE Calculate rotation angle: $ \theta = \arccos(\hat{\boldsymbol{v}}_1 \cdot \hat{\boldsymbol{v}}_2) $
\IF{$\|\boldsymbol{k}\| < 10^{-2}$}
    \RETURN Identity matrix $\boldsymbol{I}_{3\times3}$
\ENDIF
\STATE Normalize rotation axis: $ \hat{\boldsymbol{k}} = \frac{\boldsymbol{k}}{\|\boldsymbol{k}\|} $
\STATE Construct skew-symmetric matrix $\boldsymbol{K}$:
$$
\boldsymbol{K} = \begin{bmatrix}
0 & -\hat{k}_z & \hat{k}_y \\
\hat{k}_z & 0 & -\hat{k}_x \\
-\hat{k}_y & \hat{k}_x & 0
\end{bmatrix}
$$
\STATE Compute Rodrigues rotation matrix:

\quad\quad\quad\quad 
$
\boldsymbol{R} = \boldsymbol{I}_{3\times3} + \sin\theta\boldsymbol{K} + (1-\cos\theta)\boldsymbol{K}^2 
$
\RETURN $\boldsymbol{R}$
\end{algorithmic}
\end{algorithm}

\subsection{Normalized Mapping of Humans to Humanoids}
\label{sec: Normalized mapping of humans to humanoids}

Given the structural differences between humans and humanoids, generated human motion sequences often require preprocessing to align with the kinematic characteristics of robots. While fine-tuning the SMPL model \cite{loper2023smpl} for robot morphology alignment has shown promise on fixed datasets, this approach becomes labor-intensive with diverse generated data. To address this, we propose a normalization method that simplifies human motion sequences into four basic joint chains (arms and legs) for efficient mapping of human postures to robot joint states.


The generated human motion sequences are represented in world coordinates using skeleton point data obtained from the diffusion model, as shown in Fig.~\ref{fig: skeleton point}. Focusing on limb movements, we express the 3D poses of limbs in robot coordinates, including both position and rotation. The robot coordinate system is defined with the positive $x$-axis as forward, the positive $y$-axis as leftward, and the positive $z$-axis as upward.


To compute the transformation matrix between the robot and the world coordinates, we use the root forward vector $\boldsymbol{v}_\text{RF}$ derived as the cross product of the vector from the root to the right hip $\boldsymbol{v}_\text{RRH}$ and the vector from the root to the left hip $\boldsymbol{v}_\text{RLH}$. For each frame in the motion sequence, the rotation matrix is computed using the root forward vector of the first frame, denoted as $\boldsymbol{v}_\text{RF0}$, and the current root forward vector $\boldsymbol{v}_\text{RF}$, as described in Procedure~\ref{proc:rotation_matrix}.

For limb movements, we consider the left arm and left leg as example. The position of the left arm is represented by the vector from the left shoulder to the left wrist $\boldsymbol{v}_\text{LSW}$, normalized by the arm length to yield the norm-position. The rotation of the left arm is described by the vector $\boldsymbol{v}_\text{LEW}$, pointing from the left elbow to the left wrist. For the NAO robot, the default arm posture is horizontal, represented by $\boldsymbol{v}_\text{LEW0}=\left[1,0,0\right]$. The rotation matrix between $\boldsymbol{v}_\text{LEW0}$ and $\boldsymbol{v}_\text{LEW}$ is computed and converted to a quaternion, providing the left arm's rotation. Similarly, the left leg's position is represented by the vector $\boldsymbol{v}_\text{LHA}$, normalized by the leg length, and its rotation is described by $\boldsymbol{v}_\text{LAF}$, pointing from the left ankle to the left foot. The default foot direction is $\boldsymbol{v}_\text{LAF0}=\left[1,0,0\right]$, and the corresponding quaternion is derived analogously.

\begin{figure}
    \centering
    \includegraphics[width=.8\linewidth]{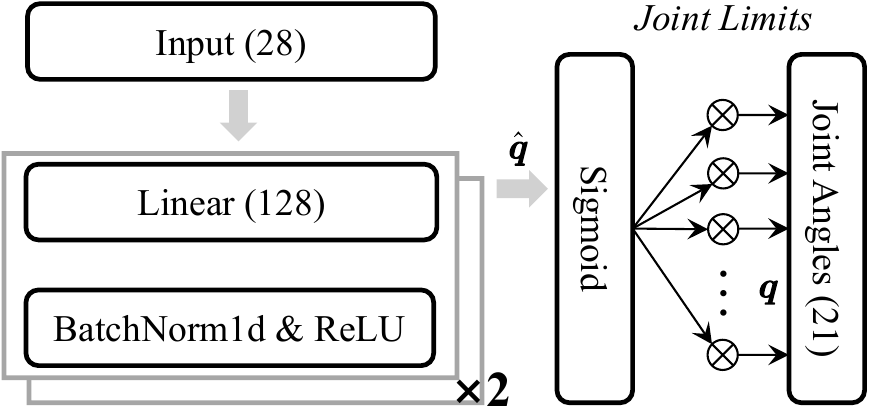}
    \caption{The architecture of the angle signal net.}
    \label{fig:angle signal net arch}
\end{figure}

The angle signal net maps input feature vectors, formed by concatenating the end-effector quaternion $\boldsymbol{Q}\in\mathbb{R}^4$ and displacement $\boldsymbol{D}\in\mathbb{R}^3$, to joint-space configurations while respecting the robot's physical joint limits as illustrated in Fig.~\ref{fig:angle signal net arch}. The network has an input dimension of $\boldsymbol{D}_\text{in}=28$ and an output dimension of $\boldsymbol{D}_\text{out}=21$, representing the predicted joint angles. It consists of two hidden layers with 128 units each, followed by batch normalization and ReLU activation. To ensure predictions remain within feasible joint limits, the raw output $\hat{\boldsymbol{q}}$ is passed through a sigmoid function:
\begin{equation}
    \boldsymbol{q}=\sigma(\hat{\boldsymbol{q}})\odot\left(\boldsymbol{q}_{\max}-\boldsymbol{q}_{\min}\right)+\boldsymbol{q}_{\min}
    \label{eq:sigmoid raw output}
\end{equation}
where $\sigma(\cdot)$ is the sigmoid function, $\odot$ denotes the element-wise multiplication, and $\boldsymbol{q}_{\max}$ and $\boldsymbol{q}_{\min}$ are the upper and lower joint limits, respectively. 

The proposed loss function, referred to as NPR Loss, evaluates discrepancies between predicted and target kinematic configurations by jointly optimizing translational and rotational accuracy. The rotational error is measured using a quaternion-based loss, which computes the angular discrepancy between predicted and target orientations:
\begin{equation}
    \begin{aligned}
        \boldsymbol{e} &= \boldsymbol{q}_\text{target}\otimes\boldsymbol{q}_\text{predict}^{-1}\\
        \mathcal{L}_\text{quat} &= 2 \cdot \arccos(w)
    \end{aligned}
\end{equation}
where $\otimes$ demotes quaternion multiplication, $\boldsymbol{q}_\text{predict}^{-1}$ is the inverse of the predicted quaternion, and $w$ is the real part of $\boldsymbol{e}$. The translational loss is computed as the mean squared error (MSE) between predicted and target positions:
\begin{equation}
    \mathcal{L}_\text{trans} = \text{MSE}(\boldsymbol{p}_\text{target},\boldsymbol{p}_\text{predicted})
\end{equation}
To account for differences in limb scale, the translational loss is weighted by limb lengths:
\begin{equation}
    \mathcal{L}_\text{trans}^\text{total} = l_\text{arm}\cdot\mathcal{L}_\text{trans}^\text{arm}+l_\text{leg}\cdot\mathcal{L}_\text{trans}^\text{leg}
\end{equation}
where $l_\text{arm}$ and $l_\text{leg}$ are the real robot's arm and leg lengths. The NPR Loss is a weighted combination of the translational and rotational losses:
\begin{equation}
    \mathcal{L} = \omega_\text{trans}\cdot\mathcal{L}_\text{trans}^\text{total}+\omega_\text{quat}\cdot\mathcal{L}_\text{quat}
\end{equation}

Fig.~\ref{fig:wave hand trajectory} illustrates the reference and the measured trajectory of the robot's arm endpoint while performing a waving motion. The reference trajectory is obtained by mapping generated motion data through the angle signal network, while the measured trajectory is captured from the physical robot. Notably, the proposed NPR Loss accurately captures the spatial features of the mapped motion, enabling precise joint sequence commands that enhance the performance of reinforcement learning.

\section{Whole Body Control Policy Training}
\label{sec: Whole Body Control Policy Training}
\subsection{State Space and Action Space}
In the whole body control policy, the state representation plays a critical role in describing the current configuration of the humanoid robot. The state includes information about the robot's joint angles, joint velocities $\dot{\boldsymbol{q}}_t$, and the global orientation and position of the robot's base. These inputs allow the control policy to understand the robot's current posture and anticipate future movements. In our approach, we encode the state as a concatenation of projected gravity $\boldsymbol{g}$, joint position targets $\boldsymbol{q}_\text{target}$, joint positions $\boldsymbol{q}_t$ and self-actions $\boldsymbol{a}_t$.

\begin{table}[!t]
    \centering
    \caption{NAO robot's joints parameters.}
    \begin{tabular}{ccccc}
    \toprule
    \textbf{Type} & \textbf{Max torque} & \textbf{Max speed} & \textbf{Stiffness} & \textbf{damping} \\ \midrule
    Head & 10.0 & 7.0 & 150.0 & 5.0 \\
    Arm & 10.0 & 7.0 & 150.0 & 5.0 \\
    Leg pitch & 20.0 & 6.4 & 200.0 & 5.0\\
    Leg roll & 20.0 & 4.0 & 150.0 & 5.0 \\
    Leg yaw pitch & 30.0 & 4.0 & 200.0 & 5.0 \\
    \bottomrule
    \end{tabular}
    \label{tab: nao joints parameters}
\end{table}

The action space defines the possible commands that the policy can output to control the robot. It consists of 21-dim joint targets. The joints are controlled by a PD controller for precise position regulation, utilizing two parameters: stiffness $K_p$ and damping $K_d$. The PD controller is formulated as:
\begin{equation}
    \tau = K_p(\hat{\boldsymbol{q}}_t-\boldsymbol{q}_t)+K_d(\hat{\dot{\boldsymbol{q}}}_t-\dot{\boldsymbol{q}}_t)+\tau_0   
\end{equation}

\begin{table}[!t]
    \centering
    \caption{Reward terms, expressions and weights}
    \begin{tabular}{ccc}
        \toprule
        \textbf{Term} & \textbf{Expression} & \textbf{Weight} \\
        \midrule
        DoF torque & $\sum{\Vert\boldsymbol{\tau}_t\Vert^2}$ & $-8.0e^{-4}$ \\
        Ankle torque & $\Vert\boldsymbol{\tau}_\text{ankle}\Vert^2$ & $-2.0e^{-3}$\\
        DoF acceleration & $\sum{\Vert\ddot{\boldsymbol{q}}_t\Vert^2}$ & $-2.5e^{-7}$ \\
        Action rate & $\sum{\Vert\boldsymbol{a}_t-\boldsymbol{a}_{t-1}\Vert}$ & $-2.0e^{-2}$ \\
        Action      & $\sum{\Vert\boldsymbol{a}_t\Vert}$ & $-6.5e^{-4}$ \\
        Torso flat & $\sum({g_{x}^2 + g_{y}^2})$ & $-1.2$\\
        Feet flat & $\sum\exp(-({a_{x,i}^2 + a_{y,i}^2})/3.046e^{-4})$ & $0.3$\\
        Undesired contacts & $\sum(f_i>10.0)$ & $-1.0$ \\
        DoF target & $\exp({-\sum{\vert\boldsymbol{q}_t-\boldsymbol{q}_\text{target}\vert}}/5.0)$ & $10.0$\\ 
        \bottomrule
    \end{tabular}
    \label{tab: reward functions}
\end{table}


The robot's actuator are driven by four types of motors. To the best of our knowledge, there are currently no publicly available simulation information for the NAO robot in IsaacSim. Therefor, we provide a set of fine-tuned joint controller parameters, where the torque is measured in \unit{\newton.\m} and speed in \unit{\radian/\second}, as shown in Table~\ref{tab: nao joints parameters}.

\subsection{Reward Formulation}
The reward function is crucial for guiding the learning process of the whole body control policy. In this work, we design a reward function that encourages the robot to accurately imitate the target human motion while maintaining stability. The reward function $r_t$ consists of several components:
\begin{enumerate}
    \item \textbf{Pose mapping reward.} Encourages the robot to align its end-effector positions (hands and feet) with the target positions generated from the text-driven human motion. Specifically, we compute this reward using the $L_2$-norm of the difference between the joint commands output by the angle signal net and the actual executed joint positions of the robot.
    \item \textbf{Security reward.} Ensure the robot does not incur self-inflicted damage during motion execution. The safety reward penalizes excessive contact forces at critical locations—specifically, the wrists, torso, and thigh links. By constraining the contact forces at these points to remain below predefined thresholds.
    \item \textbf{Joint inhibition reward.} Penalizes large or abrupt changes in joint positions, torques, velocities, and accelerations to ensure smooth and practically applicable joint actions.
\end{enumerate}

\begin{table}[!t]
    \centering
    \caption{Ranges for each parameter based on estimates of uncertainty.}
    \begin{tabular}{cc}
        \toprule
        \textbf{Parameter} & \textbf{Range} \\
        \midrule
        Static friction  & $\mathcal{U}(0.3,1.1)$ \\
        Dynamic friction & $\mathcal{U}(0.2,0.7)$ \\
        Base mass & $\mathcal{U}(0.0,0.2)$+default kg \\
        Hand mass  & $\mathcal{U}(0.0,0.1)$+default kg \\
        \midrule
        Push robot& interval$\in\mathcal{U}(4,6)$s $v_{xy}=0.2$m/s \\

        \bottomrule
    \end{tabular}
    \label{tab: domain randomization}
\end{table}

Certain components of the reward function are assigned positive weights, indicating that these actions are desirable and should be encouraged. Conversely, other components are assigned negative weights, signifying that these actions are undesirable or prohibited, and should be penalized accordingly. These defined reward functions are summarized in detail in Table \ref{tab: reward functions}.

\subsection{Termination Terms}
Termination terms define the conditions that end a training episode. These terms are essential for providing meaningful feedback to the learning process. An episode terminates if any of the following conditions are met:
\begin{enumerate}
    \item \textbf{Falling.} The robot is considered to have fallen if its base moves beyond a certain threshold from its original orientation, indicating a loss of balance. In our case, the robot has the unacceptable orientation if it rotates more than 60 degrees about the x or y axis, upon which the robot's state will terminate and reset.
    \item \textbf{Timeout Condition.}  Timeout occurs if the robot fails to complete the action within the designated time frame, resulting in the termination and reset of its state.
\end{enumerate}

\subsection{Sim-to-real Transfer}
One of the key challenges in humanoid robot control is transferring policies trained in simulation to the real world. Domain randomization involves introducing random variations in simulation parameters. Once the weights of the reward function are determined and effective motion can be trained in the simulation, domain randomization is incrementally introduced. For the NAO robot, training with domain randomization effectively reduces instability during real-world motion execution. Significantly, the joint controller of the physical NAO robot is highly sensitive to simulation parameters, and improper settings can lead to severe joint oscillations. Therefore, we chose not to randomize actuator gains. All the domain randomization we used are listed in Table~\ref{tab: domain randomization}.

\begin{figure}[!t]
    \centering
    \includegraphics[width=.8\linewidth]{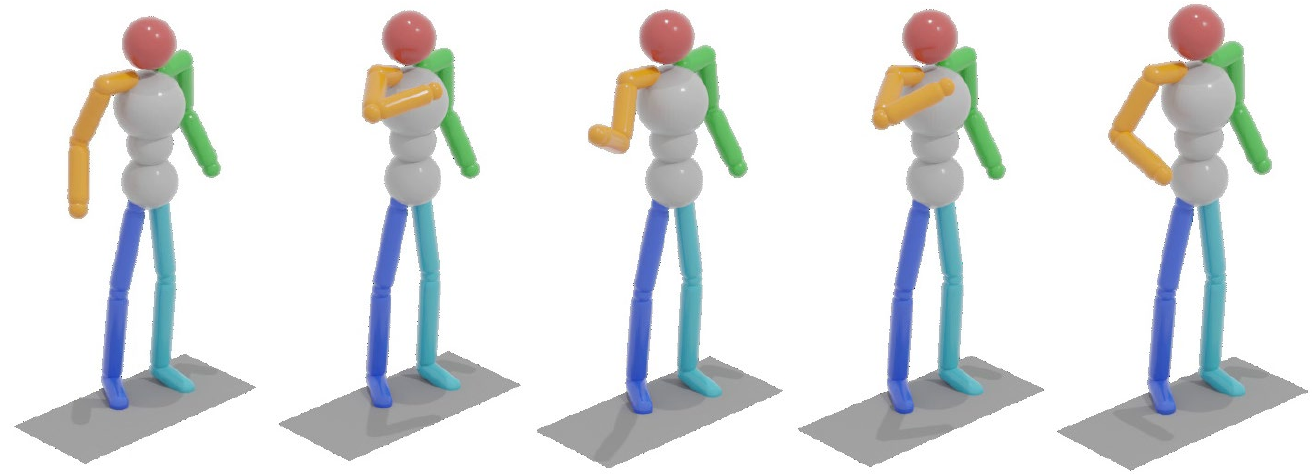}
    \caption{Human motion animation using processed outputs of the diffusion model, with the textual description ``wave right hand''.}
    \label{fig:diffusion outputs}
\end{figure}

\subsection{Training Details}
Although the angle signal network ensures the correct alignment of the HipYaw rotation, the structural design of the NAO robot introduces a unique constraint: the left and right HipYaw joints are mechanically interconnected. To further ensure accuracy during reinforcement learning, the action of the right HipYaw joint is explicitly set to mirror that of the left HipYaw joint. The policy network architecture, trained with Proximal Policy Optimization
(PPO)\cite{schulman2017proximal}, consists of a MLP with three hidden layers containing 128, 128, and 128 neurons, respectively, and exponential linear unit (ELU) as the activation function. The training was conducted on a machine equipped with an NVIDIA RTX 3080TI GPU, using IsaacLab 2.0.0 as the primary RL framework.


\section{Experimental Results}

\subsection{Simulated Experiments}

\textbf{Text-driven motion generation.} We utilize a well-trained MLCT model \cite{hu2024efficient} as motion data generator, leveraging its robust capabilities to produce high-quality motion sequences, which is trained on the HumanML3D dataset. Text prompts adhere to the annotation guidelines established by HumanML3D, employing third-person perspectives as the primary subject. The original motion data is encoded into a 263-dimensional representation and can be visualized as a human animation, as shown in Fig.~\ref{fig:diffusion outputs}. This representation includes crucial components such as 3D joint rotations, which describe the orientation of each joint in space, displacements that capture changes in joint positions, velocities that indicate the speed of these movements, and foot contact information that provides details on the interaction between the feet and the ground. To facilitate accurate orientation and visualization, we transform the encoded motion representations back into 3D joint coordinates.

\begin{figure}
    \centering
    \includegraphics[width=\linewidth]{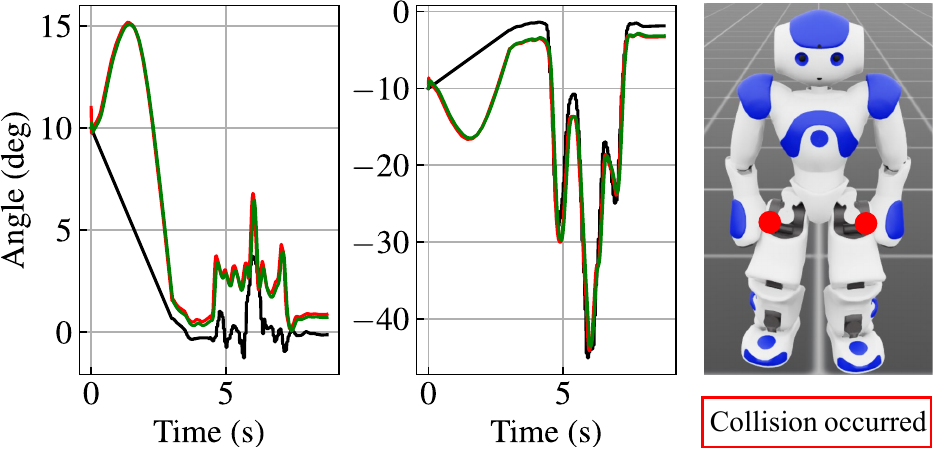}
    \caption{Joint commands (black), actions (red), and actual joint position measurements (green) for the left shoulder roll and right shoudler roll during the transition from the default pose to the reference motion. At the start of the motion, the robot adjusts its joint positions to mitigate collisions, as reflected in the significant differences between the actions and commands.}
    \label{fig: collision avoidance}
\end{figure}

\textbf{Simulated robot.} IsaacSim supports importing custom robot models. We integrate the URDF description file and STL model files obtained from the official ROS wiki into the environment and export them as USD files. We modify the STL models to address issues of unrealistic self-collisions inherent in the original files. The robot's head, torso, wrists pelvis and feet links are selected to generate collision bodies, as illustrated in Fig.~\ref{fig: nao colliders}. The NAO robot begins simulation in a crouched stance, with the hip height set to 230 mm. During the resampling phase, each robot randomly samples a single frame of joint sequences from the motion sequences as the learning target for the current episode. Additionally, we explored the use of sequential joint sequences as the learning objective for an episode, which demonstrated strong performance in simulation but encountered significant challenges when applied to the physical robot, failing to achieve effective transfer.

In the simulator, we enable self-collision detection and add contact sensors to all collision bodies of the robot. Particularly, there is a significant discrepancy between the default orientation of the robot's arms and the initial position of the reference motion. Fig.~\ref{fig: collision avoidance} illustrates the noticeable collisions that occur between the robot's hands and thighs when transitioning from the default pose to the reference motion. After optimization with the collision avoidance reward function, the robot adjusts its joint positions to prevent such collisions.

\begin{figure*}
    \centering
    \includegraphics[width=.9\textwidth]{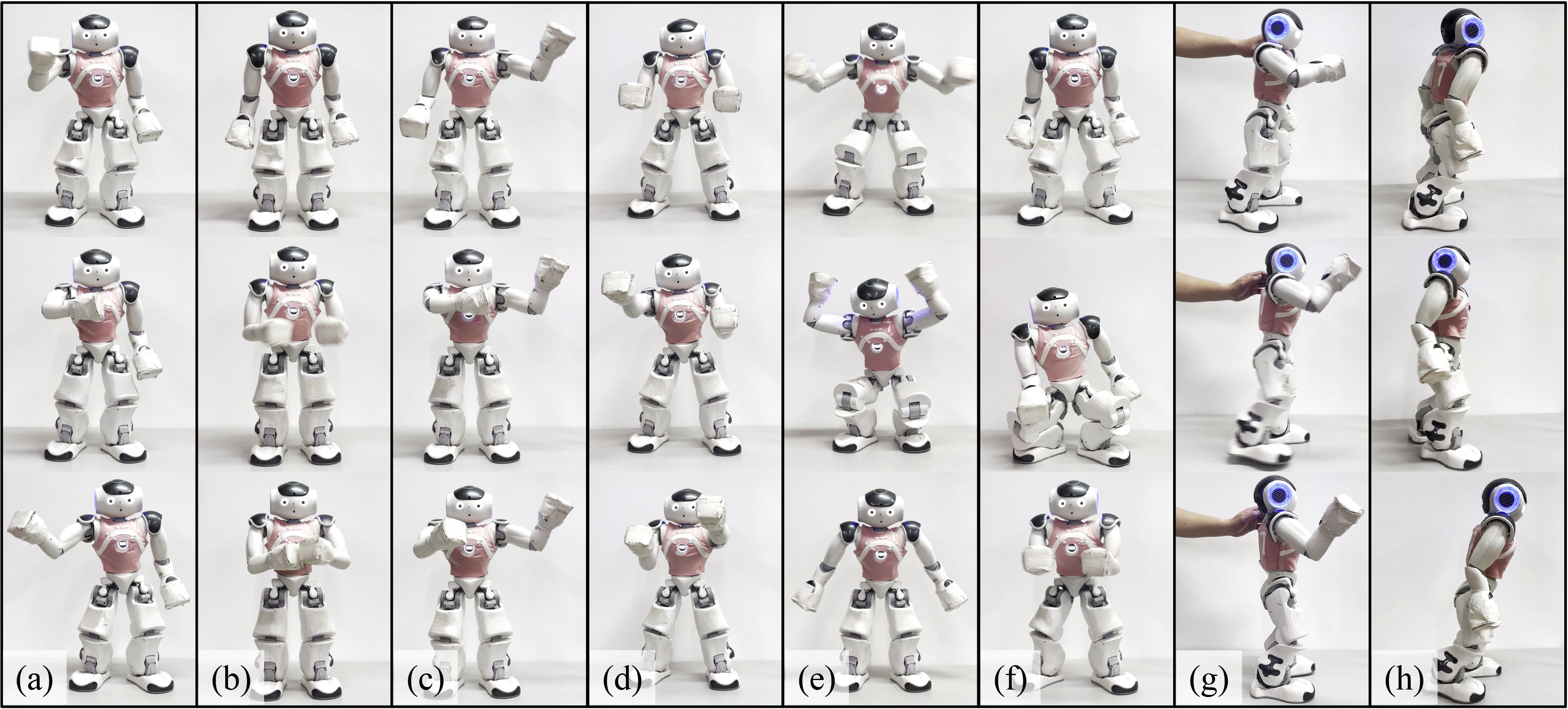}
    \caption{Time sequence snapshots of real NAO motion control, including (a) waving right hand, (b) touching the left hand with right hand, (c) playing the violin, (d) boxing, (e) squatting while raising both hands, (f) squatting to pick up an object, (g) recovering balance after being pushed, and (h) forward walking.}
    \label{fig: real world results}
\end{figure*}

\subsection{Real-world Implementaions}
\label{sub: Real-world Implementaions}
\textbf{Hardware platform.} We validated our proposed joint motion control method on the NAO robot. The NAO robot stands 57.4 cm tall and features 23 DoFs. In this work, we utilized 21 DoFs for controlling the robot's arms and legs, excluding the 2 DoFs in the head. The motion policy was exported in ONNX format and executed using ONNX Runtime for inference. The inference engine was deployed locally on NAO, with each inference taking approximately \qty{1}{\ms}. Data communication between the inference engine and the NAO robot was handled via ROS2 topics. As part of the policy input, the projected gravity vector was directly provided by the simulation environment during training. However, this cannot be directly measured on the physical robot. To address this, we employed the Madgwick Orientation Filter\cite{madgwick2010efficient} to obtain a quaternion representation of the robot's base orientation. This orientation was then applied to a unit z-axis vector to estimate the projected gravity. Other observations, such as joint positions, were gathered from the NAO robot's joint position sensors.


\textbf{Motion control results.} The real-world motion control results are shown in Fig.~\ref{fig: real world results}, where the robot demonstrates the ability to replicate upper-body motions, as shown in (a) to (e), which primarily focus on the reproduction of various upper-limb actions. These results indicate the system's capability to map human-like upper-body movements onto the robot's joint commands. For lower-body motions, we intentionally relaxed the tracking of joint commands and disregarded ankle joint commands to allow for greater flexibility in leg movements. This design choice is illustrated in Fig.~\ref{fig: real world results}(g), where the robot adjusts its balance after being pushed, demonstrating adaptive behavior rather than strict adherence to predefined joint trajectories. Additionally, Fig.~\ref{fig: real world results}(f) illustrates a limitation in the reproduction of the reference arm motion. While the intended motion was designed to position the both hands in front of the chest, only one hand remaining in front of the torso. This discrepancy arises from the reward function's emphasis on collision avoidance, as the robot's legs occupy significant space and restrict feasible arm trajectories. Finally, Fig.~\ref{fig: real world results}(h) depicts the robot's attempt at forward walking. Although the robot mimics certain characteristics of human walking, such as leg swings and a wider stance, it does not lift its feet off the ground during the motion. This incomplete replication is considered a failure case in achieving fully dynamic bipedal locomotion.


\textbf{Discussions and limitations.} To the best of our knowledge, we have thoroughly investigated the current methodologies for implementing motion control on the NAO robot using reinforcement learning policies. One notable limitation of our approach is that it samples actions from individual motion frames without accounting for temporal coherence between consecutive frames. Although we attempted to incorporate complete motion sequences as commands during reinforcement learning training, successful transfer from simulation to the physical robot has not yet been achieved.

In the case of walking motions, a failure scenario, our method does not consider the robot's velocity information. Consequently, while our approach effectively maps human motions to robot actions for stationary tasks, it struggles with dynamic motions that inherently involve self-locomotion. Furthermore, the NAO robot's limited torso degrees of freedom pose additional challenges. Specifically, motions requiring flexibility in the waist, such as bending or turning, are constrained and reduced to mappings of upper-body movements.

\section{Conclusions and Future Works}
This paper introduces a framework that enables humanoid robots to imitate human motions generated by text-driven diffusion models. The framework employs a normalized motion representation based on NPR Loss and leverages an angle signal network to map human motion data into robot joint commands. A reinforcement learning-based joint motion controller is used to optimize these commands, which are trained in simulation and successfully deployed on a physical NAO robot, demonstrating effective sim-to-real transfer. This approach enhances the flexibility of motion generation without relying on external motion capture systems and includes an NAO simulation model fully compatible with IsaacSim.

Currently, the proposed method primarily focuses on imitating joint position commands and does not account for temporal dependencies between these commands, as the current NAO simulation parameters do not support such advanced modeling. To address this limitation, further exploration will be conducted to enable the learning of continuous motion commands, thereby developing a more versatile and broadly applicable robotic simulation model.

\bibliographystyle{IEEEtran}
\bibliography{ref}

\end{document}